\documentclass{article}
\usepackage{ijcai20}

\usepackage{times}
\usepackage{soul}
\usepackage{url}
\usepackage[hidelinks]{hyperref}
\usepackage[utf8]{inputenc}
\usepackage[small]{caption}
\usepackage{graphicx}
\usepackage{amsmath}
\usepackage{amsthm}
\usepackage{booktabs}
\usepackage{algorithm}
\usepackage{algorithmic}
\usepackage{booktabs, tabularx, threeparttable, xfrac, algorithm, array, multirow, amsmath, amssymb}

\usepackage{fancyhdr}
\fancypagestyle{firststyle}
{
	
	\fancyhead[C]{Proceedings of the IJCAI-PRICAI 2020 Workshop on Explainable AI (XAI)}
}
\title{A Performance-Explainability Framework\\ to Benchmark Machine Learning Methods:\\ Application to Multivariate Time Series Classifiers}

\author{
	Kevin Fauvel
	\and
	V\'eronique Masson\And
	\'Elisa Fromont
	\affiliations
	Univ Rennes, Inria, CNRS, IRISA, France
	\emails
	kevin.fauvel@inria.fr,
	\{veronique.masson, elisa.fromont\}@irisa.fr
}

\begin{document}

\maketitle

\begin{abstract}
\vspace{-0.3em}
Our research aims to propose a new performance-explainability analytical framework to assess and benchmark machine learning methods. The framework details a set of characteristics that systematize the performance-explainability assessment of existing machine learning methods. In order to illustrate the use of the framework, we apply it to benchmark the current state-of-the-art multivariate time series classifiers.
\end{abstract}

\vspace{-1.3em}
\section{Introduction}
There has been an increasing trend in recent years to leverage machine learning methods to automate decision-making processes. However, for many applications, the adoption of such methods cannot rely solely on their prediction performance. For example, the European Union's General Data Protection Regulation, which became enforceable on 25 May 2018, introduces a right to explanation for all individuals so that they can obtain ``meaningful explanations of the logic involved'' when automated decision-making has ``legal effects'' on individuals or similarly ``significantly affecting'' them\footnote{\url{https://ec.europa.eu/info/law/law-topic/data-protection_en}}. Therefore, in addition to their prediction performance, machine learning methods have to be assessed on how they can supply their decisions with explanations.

The performance of a machine learning method can be assessed by the extent to which it correctly predicts unseen instances. A metric like the accuracy score commonly measures the performance of a classification model. However, there is no standard approach to assess explainability. 
First, there is no mathematical definition of explainability. A definition proposed by~\cite{Miller19} states that the higher the explainability of a machine learning algorithm, the easier it is for someone to comprehend why certain decisions or predictions have been made. Second, there are several methods belonging to different categories (explainability-by-design, post-hoc model-specific explainability and post-hoc model-agnostic explainability)~\cite{Du20}, which provide their own form of explanations.

The requirements for explainable machine learning methods are dependent upon the application and to whom the explanations are intended for~\cite{Tomsett18,Bohlender19}. In order to match these requirements and conduct experiments to validate the usefulness of the explanations by the end-users, there is a need to have a comprehensive assessment of the explainability of the existing methods.~\citeauthor{Doshi17} [2017] claim that creating a shared language is essential for the evaluation and comparison of machine learning methods, which is currently challenging without a set of explanation characteristics. As far as we have seen, there is no existing framework which defines a set of explanation characteristics that systematize the assessment of the explainability of existing machine learning methods.

Hence, in this paper, we propose a new framework to assess and benchmark the performance-explainability characteristics of machine learning methods. The framework hypothesizes a set of explanation characteristics, and as emphasized in~\cite{Wolf19}, focuses on what people might need to understand about machine learning methods in order to act in concert with the model outputs. The framework does not claim to be exhaustive and excludes application-specific implementation constraints like time, memory usage and privacy. It could be a basis for the development of a comprehensive assessment of the machine learning methods with regards to their performance and explainability and for the design of new machine learning methods. 
Due to space constraint, we limit the illustration of the use of the framework to one category of machine learning methods and we choose the Multivariate Time Series (MTS) classifiers. Multivariate data which integrates temporal evolution has received significant interests over the past decade, driven by automatic and high-resolution monitoring applications (e.g. healthcare~\cite{Li18}, mobility~\cite{Jiang19}, natural disasters~\cite{Fauvel20}). Moreover, the available explainability solutions to support the current state-of-the-art MTS classifiers remain limited, so this category of methods appears meaningful to assess for us.

The contributions of this paper are the following:

\begin{itemize}
	\vspace{-0.4em}
	\item We present a new performance-explainability analytical framework to assess and benchmark machine learning methods;
	\vspace{-0.3em}
	\item We detail a set of characteristics that systematize the performance-explainability assessment of existing machine learning methods;
	\vspace{-0.3em}
	\item We illustrate the use of the framework by benchmarking the current state-of-the-art MTS classifiers.
	\vspace{-0.3em}
\end{itemize}

\section{Related Work}
In this section, we first position this paper in the related work and introduce the different categories of explainability methods as a background to the notions that will be discussed in the framework. Then, we present the state-of-the-art machine learning methods that will be used to illustrate the framework, i.e. MTS classifiers.

\vspace{-0.3em}
\subsection{Explainability}
Multiple taxonomies of explainability methods have been derived from different frameworks~\cite{Guidotti2018,Ventocilla18,Du20}.
However, none of them defines a set of explanation characteristics that systematize the assessment of the explainability of existing machine learning methods. 
~\cite{Guidotti2018} provides a classification of the main problems addressed in the literature with respect to the notion of explanation and the type of machine learning systems. ~\cite{Ventocilla18} proposes a high-level taxonomy of interpretable and interactive machine learning composed of six elements (Dataset, Optimizer, Model, Predictions, Evaluator and Goodness).
And,~\cite{Du20} categorizes existing explainability methods of machine learning models into either by design or post-hoc explainability.
As our framework aims to cover all types of methods, we do not present the frameworks focusing on a particular type of explainability methods (e.g.~\cite{Lundberg17,Ancona18,Henin19}).

A five-step method to understand the requirements for explainable AI systems has been published in~\cite{Hall19}. The five steps are: explainee role definition, explanation characteristics identification, requirements collection, existing methods assessment and requirements/existing methods mapping. Our framework can be positioned as a further development of the fourth step of the method by detailing a set of explanations characteristics that systematize the assessment of existing methods. Our framework does not include application-specific implementation constraints like time, memory usage and privacy.

As a background to the notions that will be discussed in the framework, we introduce the three commonly recognized categories (explainability-by-design, post-hoc model-specific explainability and post-hoc model-agnostic explainability)~\cite{Du20} to which all of the explainability methods are belonging to.
First, some machine learning models provide explainability-by-design. These self-explanatory models incorporate explainability directly to their structures. This category includes, for example, decision trees, rule-based models and linear models.
Next, post-hoc model-specific explainability methods are specifically designed to extract explanations for a particular model. These methods usually derive explanations by examining internal model structures and parameters. For example, a method has been designed to measure the contribution of each feature in random forests~\cite{Palczewska13}; and another one has been designed to identify the regions of input data that are important for predictions in convolutional neural networks using the class-specific gradient information~\cite{Selvaraju19}.
Finally, post-hoc model-agnostic explainability methods provide explanations from any machine learning model. These methods treat the model as a black-box and does not inspect internal model parameters. For example, the permutation feature importance method~\cite{Altmann10} and the methods using an explainable surrogate model~\cite{Lakkaraju17,Lundberg17,Ribeiro18,Guidotti19} belong to this category.

The explainability methods presented reflect the diversity of explanations generated to support model predictions, therefore the need for a framework in order to benchmark the machine learning methods explainability. The next section present the MTS classifiers that will be used to illustrate the framework.

\vspace{-0.3em}
\subsection{Multivariate Time Series Classifiers}
\label{rw_MTS}
The state-of-the-art MTS classifiers consist of a diverse range of methods which can be categorized into three families: similarity-based, feature-based and deep learning methods.

Similarity-based methods make use of similarity measures to compare two MTS. 
Dynamic Time Warping (DTW) has been shown to be the best similarity measure to use along the k-Nearest Neighbors (k-NN)~\cite{Seto15}.
There are two versions of kNN-DTW for MTS: dependent (DTW$_{D}$) and independent (DTW$_{I}$). Neither dominates over the other~\cite{Shokoohi17} from an accuracy perspective but DTW$_{I}$ allows the analysis of distance differences at feature level. 

Next, feature-based methods include shapelets (gRSF~\cite{Karlsson16}, UFS~\cite{Wistuba15}) and bag-of-words (LPS~\cite{Baydogan16}, mv-ARF~\cite{Tuncel18}, SMTS~\cite{Baydogan14}, WEASEL+MUSE~\cite{Schafer17}) models. 
WEASEL+MUSE shows better results compared to gRSF, LPS, mv-ARF, SMTS and UFS on average (20 MTS datasets). WEASEL+MUSE generates a bag-of-words representation by applying various sliding windows with different sizes on each discretized dimension (Symbolic Fourier Approximation) to capture features (unigrams, bigrams, dimension identification). Following a feature selection with chi-square test, it classifies the MTS based on a logistic regression classifier.

Then, deep learning methods use Long-Short Term Memory (LSTM) and/or Convolutional Neural Networks (CNN). According to the results published, the current state-of-the-art model (MLSTM-FCN) is proposed in~\cite{Karim19} and consists of a LSTM layer and a stacked CNN layer along with Squeeze-and-Excitation blocks to generate latent features.

Therefore, we choose to benchmark the performance-explainability of the best-in-class for each similarity-based, feature-based and deep learning category (DTW$_{I}$, WEASEL+MUSE and MLSTM-FCN classifiers). The next section introduces the performance-explainability framework, which is illustrated with the benchmark of the best-in-class MTS classifiers in section~\ref{sec:results}.

\vspace{-0.5em}
\section{Performance-Explainability Framework}
The framework aims to respond to the different questions an end-user may ask to take an informed decision based on the predictions made by a machine learning model: \textit{What is the level of performance of the model? Is the model comprehensible? Is it possible to get an explanation for a particular instance? Which kind of information does the explanation provide? Can we trust the explanations? What is the target user category of the explanations?} The performance-explainability framework that we propose is composed of the following components, which will also be translated into terms specific to our application (MTS classifiers) whenever relevant:

\vspace{-0.3em}
\paragraph{Performance} \textit{What is the level of performance of the model?}
The first component of the framework characterizes the performance of a machine learning model. Different methods (e.g. holdout, k-fold cross-validation) and metrics (e.g. accuracy, F-measure, Area Under the ROC Curve) exist to evaluate the performance of a machine learning model~\cite{Witten16}. However, there is no consensus on an evaluation procedure to assess the performance of a machine learning model. Recent work suggests that the definition of such an evaluation procedure necessitates the development of a measurement theory for machine learning~\cite{Flach19}. Many of the problems stem from a limited appreciation of the importance of the \textit{scale} on which the evaluation measures are expressed.
	
Then, in current practices, the choice of a metric to evaluate the performance of a machine learning model depends on the application. According to the application, a metric aligned with the goal of the experiments is selected, which prevents the performance comparison of machine learning models across applications. 
	
Therefore, the performance component in the framework is defined as a first step towards a standard procedure to assess the performance of machine learning models. It corresponds to the relative performance of a model on a particular application. More specifically, it indicates the relative performance of the models as compared to the state-of-the-art model on a particular application and an evaluation setting. This definition allows the categorization of the models' performance on an application and an evaluation setting. In the case of different applications with a similar machine learning task, the performance component can give the list of models which outperformed current state-of-the-art models on their respective application. Thus, it points to certain models that could be interesting to evaluate on a new application, without providing guarantee that these models would perform the same on this new application. We propose an assessment of the performance in three categories:
\begin{enumerate}
	\vspace{-0.4em}
	\item[$\bullet$] \textit{Best}: best performance. It corresponds to the performance of the first ranked model on the application following an evaluation setting (models, evaluation method, datasets);
	\vspace{-0.3em}
	\item[$\bullet$] \textit{Similar}: performance similar to that of the state-of-the-art models. Based on the same evaluation setting, it corresponds to all the models which do not show a statistically significant performance difference with the second ranked model. For example, the statistical comparison of multiple classifiers on multiple datasets is usually presented on a critical difference diagram~\cite{Demsar06};
	\vspace{-1em}
	\item[$\bullet$] \textit{Below}: performance below that of the state-of-the-art models. It corresponds to the performance of the remaining models with the same evaluation setting.
\end{enumerate}
	
\paragraph{Model Comprehensibility} \textit{Is the model comprehensible?}
The model comprehensibility corresponds to the ability for the user to understand how the model works and produces certain predictions. Comprehensibility is tightly linked to the model complexity; yet, there is no consensus on model complexity assessment~\cite{Guidotti2018}. Currently, two categories of models are commonly recognized: ``white-box'' models, i.e. easy-to-understand models, and ``black-box'' models, i.e. complicated-to-understand models~\cite{Lipton16}. For example, many rule-based models and decision trees are regarded as ``white-box'' models while ensemble methods and deep learning models are ``black-box'' models. Not all rule-based models or decision trees are ``white-box'' models. Cognitive limitations of humans place restrictions on the complexity of the approximations that are understandable to humans. For example, a decision tree with a hundred levels cannot be considered as an easy-to-understand model~\cite{Lakkaraju17}.

Nevertheless, the distinction between ``white-box'' models and ``black-box'' models is clear among the machine learning methods of this paper. The state-of-the-art MTS classifiers are all ``black-box'' except one which is an easy-to-understand similarity-based approach. Therefore, due to space limitation, we propose a first assessment of the comprehensibility in two categories and we plan to further elaborate this component in future work:

\begin{enumerate}
	\item[$\bullet$] \textit{Black-Box}: ``black-box'' model, i.e. complicated-to-understand models;
	\vspace{-0.3em}
	\item[$\bullet$] \textit{White-Box}: ``white-box'' model, i.e. easy-to-understand models.
\end{enumerate}
	
\paragraph{Granularity of the Explanations} \textit{Is it possible to get an explanation for a particular instance?}
The granularity indicates the level of possible explanations. Two levels are generally distinguished: global and local~\cite{Du20}. Global explainability means that explanations concern the overall behavior of the model across the full dataset, while local explainability informs the user about a particular prediction. Some methods can provide either global or local-only explainability while other methods can provide both (e.g. decision trees). Therefore, we propose an assessment of the granularity in three categories:
\begin{enumerate}
	\item[$\bullet$] \textit{Global}: global explainability;
	\vspace{-0.3em}
	\item[$\bullet$] \textit{Local}: local explainability;
	\vspace{-0.3em}
	\item[$\bullet$] \textit{Global \& Local}: both global and local explainability.
\end{enumerate}
	
\paragraph{Information Type} \textit{Which kind of information does the explanation provide?}
The information type informs the user about the kind of information communicated. The most valuable information is close to the language of human reasoning, with causal and counterfactual rules~\cite{Pearl18}. Causal rules can tell the user that certain observed variables are the causes of specific model predictions. However, machine learning usually leverages statistical associations in the data and do not convey information about the causal relationships among the observed variables and the unobserved confounding variables. The usual statistical associations discovered by machine learning methods highly depend on the machine learning task. Therefore, we first give a generic high-level definition of the information type and then we detail and illustrate it for the application case of this paper (MTS classification). We propose a generic assessment of the information type in 3 categories from the least to the most informative:

\begin{enumerate}
	\item[$\bullet$] \textit{Importance}: the explanations reveal the relative importance of each dataset variable on predictions. The importance indicates the statistical contribution of each variable to the underlying model when making decisions;
	\vspace{-0.3em}
	\item[$\bullet$] \textit{Patterns}: the explanations provide the small conjunctions of symbols with a predefined semantic (patterns) associated with the predictions;
	\vspace{-0.3em}
	\item[$\bullet$] \textit{Causal}: the most informative category corresponds to explanations under the form of causal rules;
\end{enumerate}
	
\begin{figure}[!htpb]
	\centering
	\includegraphics[width=.95\linewidth]{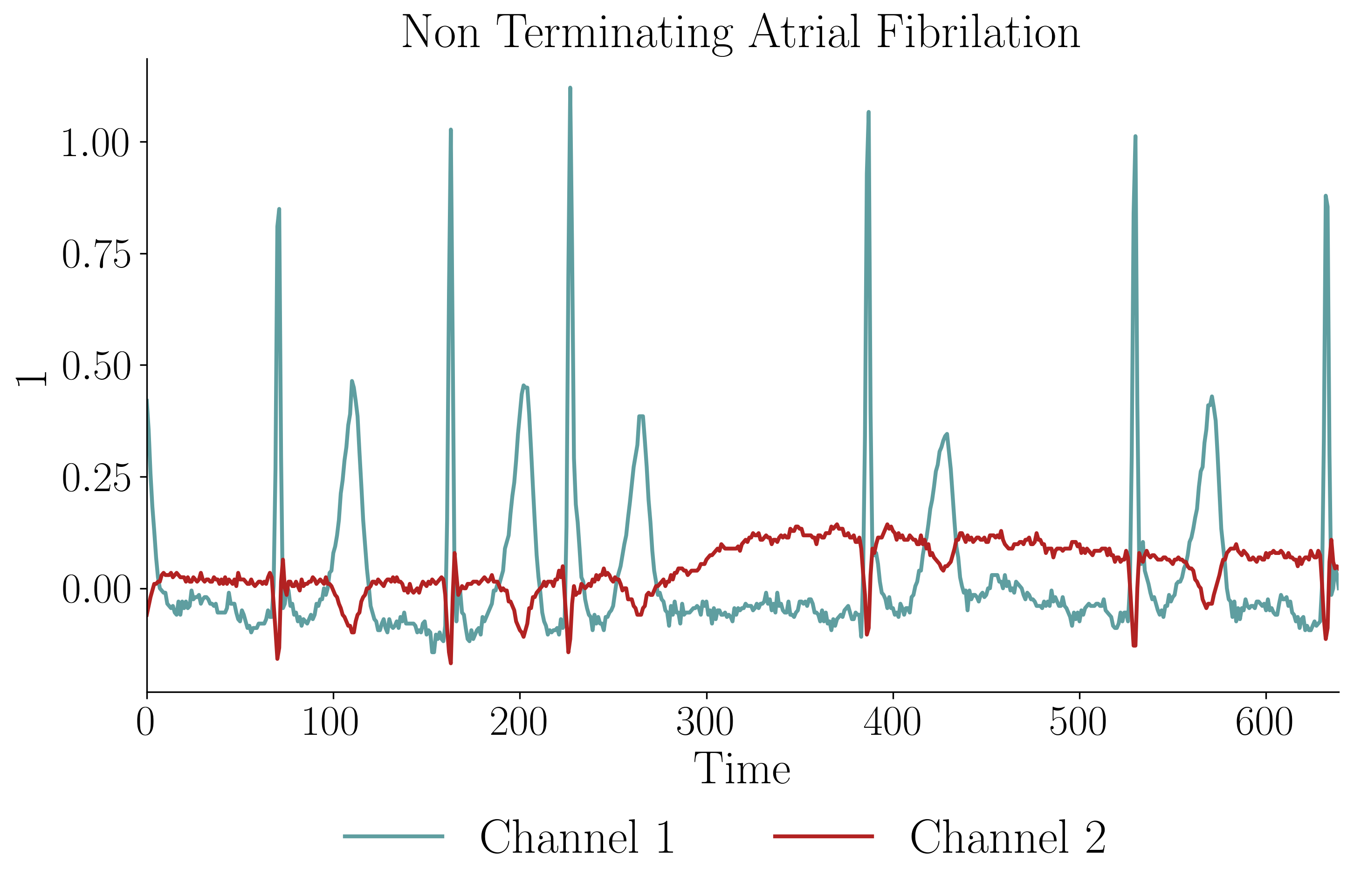}
	\caption{The first MTS sample of the UEA Atrial Fibrilation test set. It belongs to the class \textit{Non-Terminating Atrial Fibrilation} and is composed of two channels ECG on a 5 second period (128 samples per second).}
	\label{fig:framework_MTS_example}
	\vspace{-1.5em}
\end{figure}

In this paper, the issue of Multivariate Time Series (MTS) classification is addressed. A MTS $M=\{x_1,...,x_d\} \in \mathcal{R}^{d*l}$ is an ordered sequence of $d \in \mathcal{N}$ streams with $x_i=(x_{i,1},...,x_{i,l})$, where $l$ is the length of the time series and $d$ is the number of multivariate dimensions. Thus, considering the MTS data type, the information can be structured around the features, i.e. the observed variables, and the time. We propose to decompose the 3 categories presented into 8 categories. In addition, we will illustrate each of these categories with an application in the medical field. Figure~\ref{fig:framework_MTS_example} shows the first MTS of the UEA Atrial Fibrilation~\cite{Bagnall18} test set that belongs to the class \textit{Non-Terminating Atrial Fibrilation}. This MTS is composed of two dimensions (two channels ECG) with a length of 640 (5 second period with 128 samples per second). It is worth noting that the explanations provided to illustrate each category are assumptive rather than validated, they are given as illustrative in nature.

\begin{table*}[!htpb]
	\centering
	\scriptsize
	\begin{threeparttable}
		\caption{Summary of framework results of the state-of-the-art MTS classifiers.}
		\label{tab:MTS}
		\vspace{-1em}
		\begin{tabularx}{.71\linewidth}{l>{\centering}m{3cm}>{\centering}m{3cm}>{\centering\arraybackslash}m{3cm}}
			\toprule
			\textbf{} & \textbf{Similarity-Based} & \textbf{Feature-Based} & \textbf{Deep Learning}\\
			\textbf{} & DTW$_{I}$ & WEASEL+MUSE with SHAP & MLSTM-FCN with SHAP\\
			\midrule\midrule[.1em]
			Performance & Below\tnote{1} & Similar\tnote{1} & Best\tnote{1}\\
			Comprehensibility & White-Box & Black-Box & Black-Box\\
			Granularity & Local & Both Global \& Local & Both Global \& Local\\
			Information & Features+Time & Features+Time & Features+Time\\
			Faithfulness & Perfect & Imperfect & Imperfect\\
			User & Domain Expert & Domain Expert & Domain Expert\\
			\bottomrule
		\end{tabularx}
		\begin{tablenotes}
			\item[1] Predefined train/test splits and an arithmetic mean of the accuracies on 35 public datasets [Karim et al., 2019]. As presented in section~\ref{rw_MTS}, the models evaluated in the benchmark are: DTW$_{D}$, DTW$_{I}$, gRSF, LPS, MLSTM-FCN, mv-ARF, SMTS, UFS and WEASEL+MUSE.
		\end{tablenotes}
	\vspace{-1.6em}
	\end{threeparttable}
\end{table*}

\begin{enumerate}
	\item[$\bullet$] \textit{Features} (\textit{Importance}): the explanations reveal the relative importance of the features on predictions. For example, in order to support a model output from the MTS of the Figure~\ref{fig:framework_MTS_example}, the explanations could tell the user that the channel 2 has a greater importance on the prediction than the channel 1;
	\vspace{-0.3em}
	\item[$\bullet$] \textit{Features + Time} (\textit{Importance}): the explanations provide the relative importance of the features and timestamps on predictions. For example, in order to support a model output from the MTS of the Figure~\ref{fig:framework_MTS_example}, the explanations could tell the user that the channel 2 has a greater importance on the prediction than the channel 1 and that the timestamps are in increasing order of importance on the prediction;
	\vspace{-0.3em}
	\item[$\bullet$] \textit{Features + Time + Values} (\textit{Importance}): in addition to the relative importance of the features and timestamps on predictions, the explanations indicate the discriminative values of a feature for each class. For example in Figure~\ref{fig:framework_MTS_example}, the explanations could give the same explanations as the previous category, plus, it could tell the user that the timestamps with the highest importance are associated with high values (values above 0.15) on the channel 2;
	\vspace{-0.3em}
	\item[$\bullet$] \textit{Uni Itemsets} (\textit{Patterns}): the explanations provide patterns under the form of groups of values, also called itemsets, which occur per feature and are associated with the prediction. For example, in order to support a model output from the MTS of the Figure~\ref{fig:framework_MTS_example}, the explanations could tell the user that the following itemsets are associated with the prediction: \{channel 1: extremely high value (above 1); channel 1: low value (below -0.05)\} and \{channel 2: high value (above 0.15); channel 2: extremely low value (below -0.1)\}. The first itemset can be read as: the prediction is associated with the occurence on the channel 1 of an extremely high value being above 1 and a low value being below -0.05 at another moment, without information on which one appears first;
	\vspace{-0.3em}
	\item[$\bullet$]\textit{Multi Itemsets} (\textit{Patterns}): the explanations provide patterns under the form of multidimensional itemsets, i.e. groups of values composed of different features, which are associated with the prediction. For example, in order to support a model output from the MTS of the Figure~\ref{fig:framework_MTS_example}, the explanations could tell the user that the following itemset is associated with the prediction: \{channel 1: extremely high value (above 1); channel 2: high value (above 0.15)\};
	\vspace{-0.3em}
	\item[$\bullet$] \textit{Uni Sequences} (\textit{Patterns}): the explanations provide patterns under the form of ordered groups of values, also called sequences, which occur per feature and are associated with the prediction. For example, in order to support a model output from the MTS of the Figure~\ref{fig:framework_MTS_example}, the explanations could tell the user that the following sequences are associated with the prediction: $<$channel 1: extremely high value (above 1); channel 1: low value (below -0.05)$>$ and $<$channel 2: high values (above 0.15) with an increase during 1 second$>$. The first sequence can be read as: the prediction is associated with the occurrence on the channel 1 of an extremely high value being above 1 followed by a low value being below -0.05;
	\vspace{-0.3em}
	\item[$\bullet$] \textit{Multi Sequences} (\textit{Patterns}): the explanations provide patterns under the form of multidimensional sequences, i.e. ordered groups of values composed of different features, which are associated with the prediction. For example, in order to support a model output from the MTS of the Figure~\ref{fig:framework_MTS_example}, the explanations could tell the user that the following sequence is associated with the prediction: $<$channel 1: extremely high value (above 1); channel 2: high values (above 0.15) with an increase during 1 second$>$;
	\vspace{-0.3em}
	\item[$\bullet$] \textit{Causal}: the last category corresponds to explanations under the form of causal rules. For example, in order to support a model output from the MTS of the Figure~\ref{fig:framework_MTS_example}, the explanations could tell the user that the following rule applies: if (channel 1: extremely high value (above 1)) \& (channel 2: high values (above 0.15) with an increase during 1 second), then the MTS belongs to the class \textit{Non-Terminating Atrial Fibrilation}.
\end{enumerate}

\paragraph{Faithfulness} \textit{Can we trust the explanations?}
The faithfulness corresponds to the level of trust an end-user can have in the explanations of model predictions, i.e. the level of relatedness of the explanations to what the model actually computes. An explanation extracted directly from the original model is faithful by definition. Some post-hoc explanation methods propose to approximate the behavior of the original ``black-box'' model with an explainable surrogate model. The explanations from the surrogate models cannot be perfectly faithful with respect to the original model~\cite{Rudin19}. The fidelity criteria is used to quantify the faithfulness by the extent to which the surrogate model imitates the prediction score of the original model~\cite{Guidotti2018}.

In this paper, two MTS classifiers use an explainable surrogate model among the three state-of-the-art methods presented in section~\ref{sec:results}. However, there is no need to distinguish between the degree of fidelity of the surrogate models for the purpose of the comparison in this paper. Therefore, due to space limitation, we propose a first assessment of the faithfulness in two categories and we plan to further elaborate this component in future work:
\begin{enumerate}
	\item[$\bullet$] \textit{Imperfect}: imperfect faithfulness (use of an explainable surrogate model);
	\vspace{-0.3em}
	\item[$\bullet$] \textit{Perfect}: perfect faithfulness.
\end{enumerate}
	
\paragraph{User category} \textit{What is the target user category of the explanations?}
The user category indicates the audience to whom the explanations are accessible. 
The user's experience will affect what kind of \textit{cognitive chunks} they have, that is, how they organize individual elements of information into collections~\cite{Neat03}. Thus, it could be interesting to categorize the user types and associate with the model to whom the explanations will be accessible to. The broader the audience, the better are the explanations. Therefore, we propose an assessment in three categories:
\begin{enumerate}
	\item[$\bullet$] \textit{Machine Learning Expert};
	\vspace{-0.3em}
	\item[$\bullet$] \textit{Domain Expert}:  domain experts (e.g. professionals, researchers);
	\vspace{-0.3em}
	\item[$\bullet$] \textit{Broad Audience}: non-domain experts (e.g. policy makers).
	\vspace{-0.3em}
\end{enumerate}

\begin{figure*}[!htpb]
	\centering
	\includegraphics[width=0.65\linewidth]{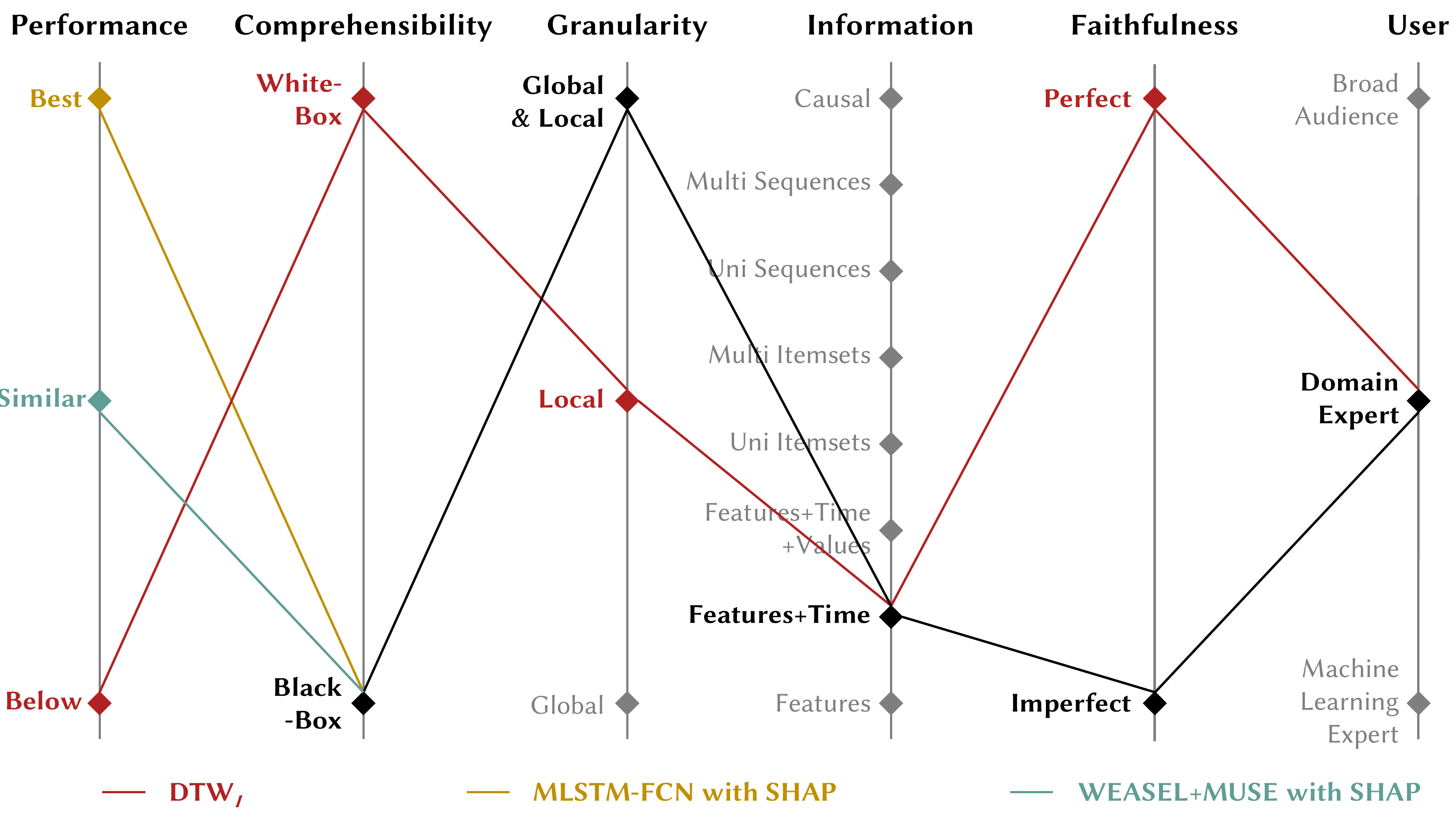}
	\vspace{-0.5em}
	\caption{Parallel coordinates plot of the state-of-the-art MTS classifiers. Performance evaluation method: predefined train/test splits and an arithmetic mean of the accuracies on 35 public datasets [Karim et al., 2019]. As presented in section~\ref{rw_MTS}, the models evaluated in the benchmark are: DTW$_{D}$, DTW$_{I}$, gRSF, LPS, MLSTM-FCN, mv-ARF, SMTS, UFS and WEASEL+MUSE.}
	\label{fig:coordinates_example}
	\vspace{-1.3em}
\end{figure*}

In order to compare the methods visually using the proposed framework, the different aspects can be represented on a parallel coordinates plot. A parallel coordinate plot allows a 2-dimensional visualization of a high dimensional dataset and is suited for the categorical data of this framework. The next section presents an example of parallel coordinates plots comparing the state-of-the-art MTS classifiers.

\vspace{-0.7em}
\section{Application to Multivariate Time Series Classifiers}
\label{sec:results}
This section shows how the framework presented in the previous section can be used to assess and benchmark the state-of-the-art MTS classifiers. As introduced in section~\ref{rw_MTS}, the state-of-the-art MTS classifiers are: DTW$_{I}$, MLSTM-FCN and WEASEL+MUSE. The results of the assessment are summarized in Table~\ref{tab:MTS}, illustrated in Figure~\ref{fig:coordinates_example} and detailed in the following paragraphs.

The first MTS classifier belongs to the similarity-based category and is the one-nearest neighbor MTS classifier with DTW distance (DTW$_{I}$). DTW$_{I}$ classifies MTS based on the label of the nearest sample and a similarity calculated as the cumulative distances of all dimensions independently measured under DTW. For each MTS, the explanation supporting the classification is the ranking of features and timestamps in decreasing order of their DTW distance with the nearest MTS.
Based on predefined train/test splits and an arithmetic mean of the accuracies, DTW$_{I}$ underperforms the current state-of-the-art MTS classifiers on the 35 public datasets (Performance: \textit{Below}). The results from~\cite{Karim19} shows that DTW$_{I}$ has a statistically significant lower performance than MLSTM-FCN and WEASEL+MUSE. Furthermore, DTW$_{I}$ supports its predictions with limited information (Information: \textit{Features+Time}) that needs to be analyzed by a domain expert to ensure that it is relevant for the application (User: \textit{Domain Expert}). However, DTW$_{I}$ is an easy-to-understand model (Comprehensibility: \textit{White-Box}) which provides faithful explanations (Faithfulness: \textit{Perfect}) for each MTS (Granularity: \textit{Local}). 

Then, we can analyze MLSTM-FCN and WEASEL+ MUSE together. First, based on predefined train/test splits and an arithmetic mean of the accuracies, MLSTM-FCN exhibits the best performance on the 35 public datasets (Performance: \textit{Best})~\cite{Karim19}, followed by WEASEL+MUSE (Performance: \textit{Similar}). Second, both MLSTM-FCN and WEASEL+MUSE are ``black-box'' classifiers without being explainable-by-design or having a post-hoc model-specific explainability method. Thus, the explainability characteristics of these models depend on the choice of the post-hoc model-agnostic explainability method. We have selected SHapley Additive exPlanations (SHAP)~\cite{Lundberg17}, a state-of-the-art post-hoc model-agnostic explainability method offering explanations at all granularity levels. SHAP method measures how much each variable (Features+Time) impacts predictions and comes up with a ranking of the variables which could be exploited by domain experts. The combination of MLSTM-FCN and WEASEL+MUSE with SHAP enables them to outperform DTW$_{I}$ while reaching explanations with a similar level of information (Information: \textit{Features+Time}, DTW$_{I}$: \textit{Features+Time}), in the meantime remaining accessible to the same user category (User: \textit{Domain Expert}, DTW$_{I}$: \textit{Domain Expert}). However, as opposed to DTW$_{I}$, SHAP relies on a surrogate model which cannot provide perfectly faithful explanations (Faithfulness: \textit{Imperfect}, DTW$_{I}$: \textit{Perfect}).

Therefore, based on the performance-explainability framework introduced, if a ``white-box'' model and perfect faithfulness are not required, it would be preferable to choose MLSTM-FCN with SHAP instead of the other state-of-the-art MTS classifiers on average on the 35 public datasets. In addition to its better level of performance, MLSTM-FCN with SHAP provides the same level of information and at all granularity levels.

However, the imperfect faithfulness of the explanations could prevent the use of MLSTM-FCN with a surrogate explainable model on numerous applications. In addition, the level of information provided to support the predictions remains limited (Information: \textit{Features+Time}). Therefore, based on the assessment of the current state-of-the-art MTS classifiers with the framework proposed, it would be valuable for instance to design some new high-performing MTS classifiers which provide faithful and more informative explanations. For example, it could be interesting to work in the direction proposed in~\cite{Fauvel20LCE}. It presents a new MTS classifier (XEM) which reconciles performance (Performance: \textit{Best}) and faithfulness while providing the time window used to classify the whole MTS (Information: \textit{Uni Sequences}). XEM is based on a new hybrid ensemble method that combines an explicit approach to handle the bias-variance trade-off and an implicit approach to individualize classifier errors on different parts of the training data~\cite{Fauvel19}. 
Nevertheless, the explanations provided by XEM are only available per MTS (Granularity: \textit{Local}) and the level of information could be further improved. As suggested by the authors, it could be interesting to analyze the time windows identified for each class to determine if they contain some common multidimensional sequences (Information: \textit{Multi Sequences}, Granularity: \textit{Both Global \& Local}). These patterns could also broaden the audience as they would summarize the key information in the discriminative time windows.

\vspace{-0.7em}
\section{Conclusion}
We have presented a new performance-explainability analytical framework to assess and benchmark the machine learning methods. The framework details a set of characteristics that systematize the performance-explainability assessment of machine learning methods. In addition, it can be employed to identify ways to improve current machine learning methods and to design new ones. Finally, we have illustrated the use of the framework by benchmarking the current state-of-the-art MTS classifiers.
With regards to future work, we plan to further elaborate the definition of the different components of the framework (especially the \textit{Model Comprehensibility}, the \textit{Information Type} and the \textit{Faithfulness}) and evaluate the relevance of integrating new components. Then, we plan to apply the framework extensively to assess the different types of existing machine learning methods.

\section*{Acknowledgments}
This work was supported by the French National Research Agency under the Investments for the Future Program (ANR-16-CONV-0004) and the Inria Project Lab ``Hybrid Approaches for Interpretable AI'' (HyAIAI).

\bibliographystyle{named}
\bibliography{ijcai20}

\end{document}